\documentclass[journal]{IEEEtran}

\usepackage{url}

\usepackage{times}
\usepackage{epsfig}
\usepackage{graphicx}
\usepackage{amsmath}
\usepackage{amssymb}
\usepackage{algorithm,algpseudocode}

\usepackage{bm}
\usepackage{caption}

\usepackage{url}
\usepackage{booktabs}
\usepackage{multirow}
\usepackage{multirow}
\usepackage{wrapfig}
\usepackage{wrapfig,lipsum,booktabs}
\usepackage{graphicx}
\usepackage{array}
\usepackage{color, colortbl}
\usepackage{array,hhline}
\usepackage[]{hyperref}
\usepackage{soul}

\usepackage{color, colortbl}

\newif\ifshowfig\showfigtrue

\ifCLASSINFOpdf
\else
\fi
\begin{document}


\title{Differential Encoding for Improved Representation Learning over Graphs}



  
  \author{Haimin~Zhang, Jiaohao~Xia,  Min~Xu,~\IEEEmembership{Member,~IEEE}

 \thanks{Haimin Zhang, Jiahao Xia and Min Xu (\emph{corresponding author})  are with the School of Electrical and Data Engineering, Faculty of Engineering and Information Technology, University of Technology Sydney, 15 Broadway, Ultimo, NSW 2007, Australia (Emails: Haimin.Zhang@uts.edu.au, Jiahao.Xia@student.uts.edu.au, Min.Xu@uts.edu.au).}

}  
\maketitle
\thispagestyle{empty}


\begin{abstract}
Combining the message-passing paradigm with the global attention mechanism   has emerged as an effective framework for learning over graphs.
The  message-passing paradigm   and the global attention mechanism fundamentally generate node embeddings  based on information aggregated  from a node's local neighborhood or from the whole graph. 
The most basic and commonly used aggregation approach  is to take the sum of information from a node's local neighbourhood or from  the whole graph.
However, it is unknown if the dominant information is from a node itself or from the node’s neighbours (or the rest of the graph nodes). Therefore, there  exists information lost at each layer of embedding generation, and this information lost could be accumulated and become more serious when more layers  are used in the model. 
In this paper, we present a differential encoding method to address the issue of information lost.
The idea of our method is to encode the differential representation between the information  from a node's neighbours (or the rest of the graph nodes)  and that from the node itself. 
The obtained differential encoding is then combined with the original aggregated local or global representation to generate the updated node embedding.
By integrating  differential encodings, the representational ability of generated node embeddings is improved.
The differential encoding method is empirically evaluated on different graph tasks on seven benchmark datasets.
The results show that it is a general method that improves the message-passing update and the global attention update, advancing the state-of-the-art  performance for graph representation learning on these datasets.

\end{abstract}


\IEEEpeerreviewmaketitle

\section{Introduction}
Graphs are a foundational data structure that can be used to represent data in a wide range of domains.  
Molecules, protein-protein interaction networks, social networks and citation networks---all these type of data can be represented using graphs.
It is significant to develop general models that are able to learn and generalize from the graph-structured data. 
Recent years have seen increasing studies devoted to learning from graph-structured data, including deep graph embedding \cite{chen2020simple,lee2023towards},  generalizing the Transformer architecture to graphs \cite{shirzad2023exphormer,kreuzer2021rethinking}, and graph normalization and regularization techniques \cite{zhao2019pairnorm,zhang2024randalign}.
These efforts have produced new state-of-the-art or human-level results in various areas such as molecular property prediction, recommendation systems and social network analysis.

A graph basically contains a set of nodes together with a set of edges between pairs of these nodes.
For example, in a molecular structure, we can use nodes to represent atoms and use edges to represent the bonds between adjacent atoms.
Unlike images and sequence data, graph-structured data have an underlying structure that is in non-Euclidean spaces.
A node can have an arbitrary number of neighbours, and there is no predefined ordering of these nodes \cite{zhang2022ssfg}. 
It is a complicated task to learn over  graphs due to the underlying non-Euclidean structure.

Early methods for graphs representation learning  primarily involves a recurrent process that iteratively
propagates node features until the node features reach an equilibrium state \cite{frasconi1998general,sperduti1997supervised}.
In recent years,  graph neural networks have become  the dominant approach for learning over graphs.
At the core, graph neural networks utilize a paradigm of message passing that generate node embeddings by aggregating information from a node's local neighbourhood. 
However, the message-passing paradigm  inherently has major drawbacks. 
Theoretically, the message-passing paradigm is  connected to the Weisfeiler-Lehman (WL) isomorphism test as well as to simple graph convolutions \cite{hamilton2020graph}, which induces bounds on the expressiveness of these graph neural network models based on the theoretical constructs.
Empirically, studies continually find that massage-passing graph neural networks suffer from the problem of over-smoothing, which can be viewed as a consequence of the neighborhood aggregation operation. 

To generate expressive graph representations, it is important to generate embeddings for nodes that depend on the graph structure and the features associated with the nodes.
However, with the over-smoothing problem, graph neural networks tend to generate embeddings for all graph nodes that are very similar to one another after several layers of message passing, resulting in node-specific information being lost.
This prevents us from building deeper graph neural networks to capture long-term dependencies between the nodes in a graph. 
An approach to alleviating this issue is to combine a global attention update in each layer of message passing \cite{rampavsek2022recipe}.
The global attention update employs an attention function that aggregates features from all nodes in a graph, enabling the model to capture information from far reaches of the graph. 

At the core of the message-passing update and the attention update, an aggregation function is utilized to aggregate information from a node's local neighbourhood or from all nodes in a graph. 
The most commonly used aggregation method simply takes the sum of the information from a node's neighbours or from  all graph nodes. 
While this aggregation method is effective and has become the dominant approach in graph neural networks and Transformer models, it is unknown if the dominant information is from a node itself or from its neighbours (or the rest of the graph nodes).
Therefore there exists information lost at each layer of embedding generation, and this information lost could be accumulated and  become serious at deeper layers.
Consequently, the representational ability of generated embeddings is reduced.
To the best of our knowledge, this issue of  information lost in the aggregation process has not been addressed in existing research.

\begin{figure}[t!]
\centering
\includegraphics[width=.38\textwidth]{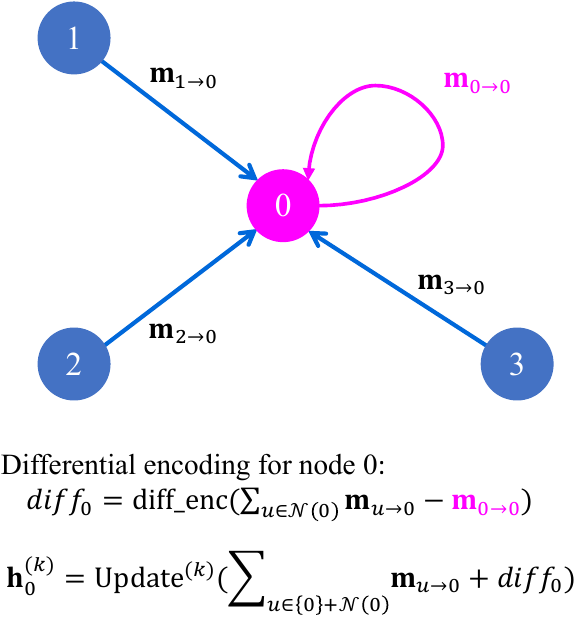}
\caption{The differential encoding encodes the difference between the message from a node's neighbours and that from the node itself. It is combined with the aggregated message at the node to generate the updated node embedding.}
\label{fig:intro}
\end{figure}

It is a common practice to encode  differential representations (or called residual representations) in feature encoding methods.
For example, the VLAD (vector of locally aggregated descriptors)  method \cite{jegou2010aggregating} encodes the residual representations between local features and vectors in a codebook for image recognition.
Inspired by the differential representation encoding approach, we propose a differential encoding method to address the information lost issue in the current aggregation method in graph neural networks and Transformer models.
Instead of simply aggregating  information from a local neighbourhood (or from all graph nodes) using summation, we compute the differential representation between the information from a node's neighbours (or from the rest of the nodes) and that from the node itself and encode the differential representation using a neural network.
The obtained encoding is added to the original aggregated representation to generate the updated embedding for the node.
Through this way, the information lost in the aggregation process is prevented.

Figure 1 demonstrates the idea of the proposed differential encoding method.
The differential encoding method  can be integrated into  the message-passing update and the global attention update.
We conduct extensive experiments on four graph tasks, i.e., graph classification, node classification, link prediction and multi-label graph classification, on seven popular benchmark datasets.
We show that the use of our differential encoding method consistently yields improved performance of message-passing graph neural networks and the Transformer model, resulting in new state-of-the-art results for representation learning on these datasets. 

To summarize, this paper provides the following contributions.
\begin{itemize}

  \item This paper proposes a differential encoding method  to address the issue of information lost in the current aggregation approach in the message-passing paradigm and  attention mechanism.
  Through reducing the information lost, the embedding capability of the message-passing update and attention update is improved.

  \item We show that our differential encoding method is a general method that  improves the message-passing update and global attention update, advancing the state-of-the-art results for graph representation learning on four graph tasks on seven benchmark datasets.


\end{itemize}

\iftrue
\section{Related Work} \label{sec:related_work}

\subsection{Graph Neural Networks}
Graph neural networks are a general framework for learning over graphs.
Most current graph neural networks can be categorized into spectral approaches and spatial approaches \cite{velickovic2018graph}.
The spectral approaches are motivated by the spectral graph theory.
The defining feature of spectral graph neural networks  is that  convolutions are defined in the spectral domain through an extension of the Fourier transform to graphs.
Bruna et al. \cite{bruna2014spectral} developed the basic graph convolutional network model, in which convolutions are defined  based on the eigendecomposition of the graph Laplacian matrix.
The ChebNet model \cite{defferrard2016convolutional} constructs convolutions according to the Chebyshev expansion of the graph Laplacian, which eliminates the process for graph Laplacian decomposition and  results in spatially closed kernels.
Kipf et al. extended  the concept of graph
convolutions to define the popular GCN model  \cite{kipf2017semi}, a layered architecture based on the first-order approximation of the spectral convolutions on graphs.

Unlike spectral graph neural networks, spatial graph neural networks define convolutions in spatially localized neighbourhoods.
The behaviour of the convolutions is analogous to that of  kernels in convolutional neural networks which aggregate features from spatially-defined patches in an image.
The GraphSAGE model \cite{hamilton2017inductive} generates the embedding for every node by sampling a fixed-size set of neighbours and aggregating  features from the sampled neighbours using an aggregator, such as the pooling aggregator and the LSTM aggregator.
The graph attention network (GAT) model \cite{velickovic2018graph}  employs the self-attention mechanism to learn a weight for each neighbour in  aggregating features from a node's local neighbourhhod, enabling the model to focus on important information in the aggregation process.
Bresson et al. \cite{bresson2017residual} introduced residual gated graph convnets (GatedGCN), which integrates edge gates, residual connections \cite{he2016deep} and batch normalization \cite{ioffe2015batch} into the graph  neural network model.

Fundamentally, both spectral and spatial graph neural networks are message-passing neural networks (MPNNs) which use a paradigm of message passing that generates embeddings through propagating messages between adjacent nodes \cite{gilmer2017neural}.
Empirically, studies  continually find that message-passing graph neural networks  suffer from the problem of over-smoothing, which can be attributed to the neighbourhood aggregation operation in the message-passing update \cite{hamilton2020graph}. 
This issue  of  over-smoothing prevents the model from capturing long-term dependencies between graph nodes.
Chen et al. \cite{chen2020simple} proposed the GCNII model which extends the basic GCN  with initial residual connection and identity mapping.
The GCNII model helps allievate the issue of over-smoothing and prevents the model performance from becoming considerably reduced using deeper layers.
Zhang et al. \cite{zhang2022ssfg} proposed a method that stochastically scale features and gradients (SSFG) during the training procedure for regularizing graph neural networks.
The SSFG method prevents over-smoothing by breaking the norm of generated embeddings to converge to  similar values.
Roth et al. \cite{roth2024rank} showed  that rank collapse of node representations is the underlying cause of over-smoothing  and introduced   the use of  the sum of Kronecker products to prevents rank collapse in graph neural networks.



\subsection{Graph Transformers}
Recent years have seen a surge in research to extend the Transformer architecture for graph-structured data.
This trend is primarily motivated by the considerable success of Transformers in various sequence learning and vision tasks.
Graphormer \cite{ying2021transformers} extends the Transformer by integrating three structural encoding methods, i.e., centrality encoding, spatial encoding and edge encoding, showing competitive performance for graph-level prediction tasks.
 GraphTrans  \cite{wu2021representing} applies a Transformer subnetwork on top of a standard graph neural network.
This  architecture outperformed the methods that explicitly encode the graph structural information on graph classification tasks.
The spectral attention network (SAN) \cite{kreuzer2021rethinking}  uses the graph Laplacian spectrum to learn  positional encodings for graph nodes.
It was the first pure Transformer-based architecture that performed well compared to message-passing graph neural networks.
Chromatic graph Transformer \cite{menegaux2023self}, which employs graph structural information and edge features, is another Transformer-based  architecture that  bypassed the paradigm of message passing.
Exphormer  \cite{shirzad2023exphormer} is a sparse graph transformer architecture that approximates the full attention mechanism using  a small number of layers, achieving the computational cost that scales linearly with the graph's size. 
GraphGPS \cite{rampavsek2022recipe} is a framework that incorporates local message-passing, global attention and  positional and structural encoding for graph representation learning.
It is a modular framework and can be scalable to  large graphs.

\iftrue
\section{Methodology}

In this section, we first briefly introduce the notations, the message-passing framework and the attention mechanism.
Then we revisit the current aggregation method in the message-passing framework and attention mechanism and present the differential encoding method to improve the  aggregation method.
Finally, we present the model architecture for representation learning over graphs.

\subsection{Preliminaries}
\textbf{Notations.}
A graph $\mathcal{G} = (\mathcal{V}, \mathcal{E})$ is defined through a set of nodes $\mathcal{V}$ and a set of edges  $\mathcal{E}$ between pairs of these nodes.
An edge  from node $u\in{V}$ to node $v\in{V}$ is denoted as $(u,v)$.
$\mathcal{N}(u)$ denotes the set  of node $u$'s neighbouring nodes.
The adjacent matrix of  $\mathcal{G}$ is denoted as $\mathbf{A} \in \mathbb{R}^{|\mathcal{V}| \times |\mathcal{V}|}$, in which $\mathbf{A}_{u,v}=1$ if $(u,v)\in {E}$ or $\mathbf{A}_{u,v}=0$ otherwise.
The degree matrix $\mathbf{D}$ of $\mathcal{G}$ is a $|\mathcal{V}| \times |\mathcal{V}|$ diagonal matrix wherein $\mathbf{D}_{ii}=\sum_j \mathbf{A}_{ij}$.
The node feature (or called attribute) associated with each node $u\in \mathcal{V}$ is denoted as $\mathbf{x}_u$.

\textbf{The Message-Passing Paradigm.}
The message-passing paradigm is at the heart of current graph neural networks.
At each layer of message passing,  a hidden embedding $\mathbf{h}_u^{(k)}$ for each node $u\in \mathcal{V}$ is generated based on the information aggregated from $u$'s local neighbourhood $\mathcal{N}(u)$.
This message-passing paradigm can be  expressed as follows: 

\begin{equation} \label{eq:message_passing}
    \begin{aligned}
    \mathbf{h}_{u}^{(k)} &=  \text{Update}^{(k)}(\text{Aggregate}^{(k)}( \\
    & \hspace{50pt}\{ {\mathbf{h}_u^{(k-1)}} \} \cup \{{\mathbf{h}_v^{(k-1)}}, \forall v \in \mathcal{N}(u) \})),  \\
    &= \text{Update}^{(k)}(\mathbf{m}_{\mathcal{N}(u)+\{u\}})
     \end{aligned}
\end{equation}
where $\text{Update}^{(k)}$ and $\text{Aggregate}^{(k)}$  are  neural networks.
\iftrue
The superscripts are used for differentiating the embeddings and functions at different layers of message passing.
\fi
During the message-passing iteration $k$, the $\text{Aggregate}$ function aggregates the embeddings of nodes in $u$'s local neighbourhood and generates an aggregated message $\mathbf{m}_{\mathcal{N}(u)+\{u\}}$. 
The $\text{Update}$ function then generates the updated embedding $\mathbf{h}_{u}^{(k)}$ according to the aggregated message.
The embeddings at $k=0$ are initialized to the node-level features, i.e., $\mathbf{h}_u^{(0)}=\mathbf{x}_u, \forall u\in \mathcal{V}$.
After $K$ message-passing iterations, every node embedding contains information from its $K$-hop neighborhood.

\textbf{The Attention Mechanism.}
The attention mechanism is a key component in Transformer-based architectures which have significantly advanced the areas of computer vision, natural language processing and beyond. 
At its core, the attention mechanism employs an attention function that maps a query and a set of key-value pairs to an output,
where the query, keys, values, and output are all vectors \cite{vaswani2017attention}.
The attention function computes the dot products of the query with keys and  uses a $\text{Softmax}$ function to generate the weights on the values.
The output is computed as a weighted sum of the values as follows: 

\begin{equation} 
    \begin{aligned}
    \mathbf{o}_i= \text{Attention}(\mathbf{q}_i,\mathbf{K},\mathbf{V}) = \text{Softmax}(\frac{\mathbf{q}_i\mathbf{K}^{T}}{\sqrt{d_q}})\mathbf{V},
    \end{aligned}
\end{equation}
where $\mathbf{q}_i$ represents the query,  $\mathbf{K}$ and $\mathbf{V}$ are matrices that contain the set of keys and the set of values respectively, and $d_q$ is the dimension of queries and keys.
In practice, the Attention function is applied simultaneously to to a set of queries which are packed into a matrix $\mathbf{Q}$. 
Rather than using a single attention function, a multi-head attention (MHA) approach is commonly employed.

\subsection{Revisiting Aggregation in the Message-Passing Update and Attention Update}
At the heart,  both the message-passing update and the attention update generate representations based on information aggregated from either a local or global set of the input.
In the message-passing framework, the aggregation function aggregates information from a node's local neighbourhood to generate an aggregated message. 
The aggregated message $\mathbf{m}_{\mathcal{N}(u)+\{u\}}$ corresponding to each node $u \in \mathcal{V}$ can be described as the sum of the message from node $u$ itself and that from its neighbours $\mathcal{N}(u)$.
Then Eq. (\ref{eq:message_passing}) can be rewritten as follows:

\begin{equation} 
    \begin{aligned}
    \mathbf{h}_{u}^{(k)}  &= \text{Update}^{(k)}(\mathbf{m}_{\mathcal{N}(u)+\{u\}})\\ 
    &= \text{Update}^{(k)}(\mathbf{m}^{(k)}_{u \to u} + \sum_{v \in \mathcal{N}(u)} \mathbf{m}_{v\to u}^{(k)}) \\ 
    \end{aligned}
\end{equation}
\normalsize
where $\mathbf{m}_{v\to u}^{(k)}$ is the message aggregated from node $v$ at node $u$.
The message $\mathbf{m}_{v\to u}^{(k)}$ can be defined using a differentiable function on $\mathbf{h}_v^{(k-1)}$ and $\mathbf{h}_u^{(k-1)}$ as well as the edge feature of $(v,u)$ if present.


In the attention mechanism,  the aggregation function (i.e., the attention function) generates a representation $\mathbf{o}_u$ for every node $u\in \mathcal{V}$ by aggregating the  value vectors packed in $\mathbf{V}$.
The representation $\mathbf{o}_u$ can also be described as  the summation of the information from  node $u$ itself and that from the rest of the nodes:

\begin{equation} 
    \begin{aligned}
    \mathbf{o}_u &= \text{Softmax}(\frac{\mathbf{q}_u\mathbf{K}^{T}}{\sqrt{d_q}})\mathbf{V} \\
               &= \sum_{u\in \mathcal{V}} a_{u} \mathbf{ V}_{\cdot u}\\
    &= a_u \mathbf{V}_{\cdot u} + \sum_{u\in V- \{u\}}a_{i}  \mathbf{V}_{\cdot u}
    \end{aligned}
\end{equation}
where $a_u$ is the attention weight on $\mathbf{V}_{\cdot u}$, i.e., the $i$-th value vector.
Because the attention weights are generated using the $\text{Softmax}$ function, the representations generated by the aggregation function are basically a convex combination of value vectors.
As compared to message-passing update, the attention mechanism aggregates information from the global set of the input, therefore it can be seen as a specific form of message passing  applied to a fully connected graph.


\begin{figure*}[t!]
\centering
\includegraphics[width=.60\textwidth]{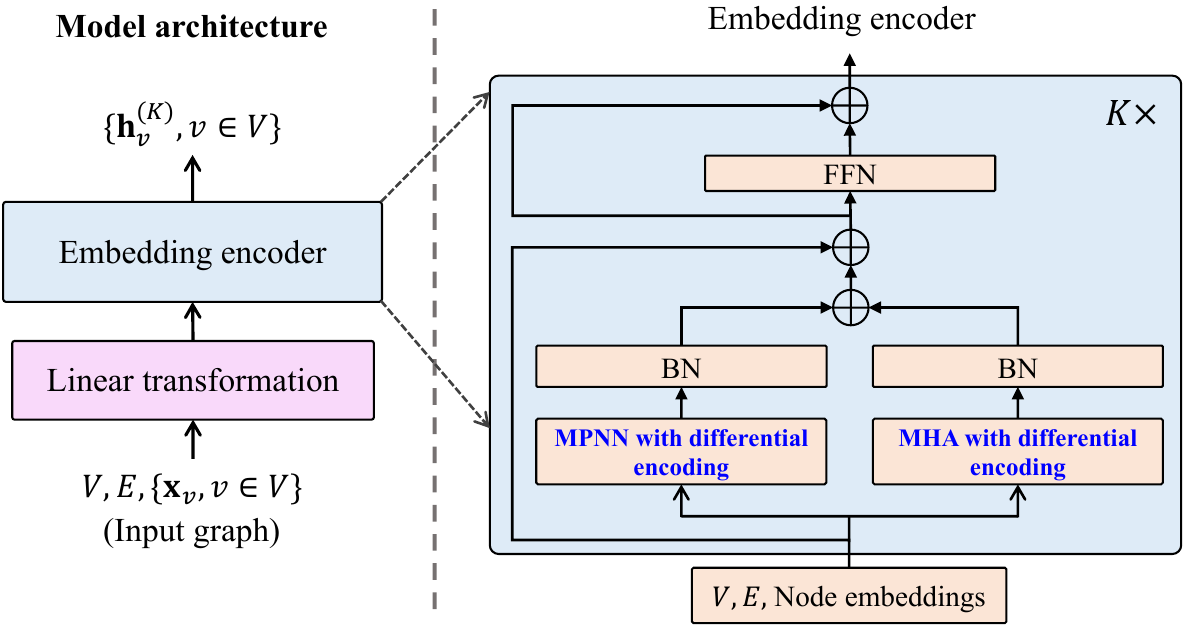}
\caption{An overview of our model architecture for learning over graphs. A layer of the embedding encoder  uses an MPNN layer to aggregate local neighbourhood features and a multi-head attention layer to aggregate  global features. The differential encoding method is applied in both the MPNN layer and the multi-head self-attention layer to improve the  representational capability of generated embeddings.}
\label{fig:overview}
\end{figure*}

\subsection{ Differential  Encoding in Aggregation}
From the above, we see that both the message-passing update and attention update use an aggregation function to aggregate information from a local set and the global set of the input, respectively.
The aggregation function basically takes the sum of information from a node itself and that from an aggregation set. 
The representation aggregated at node $u$ can be described as follows:

\begin{equation} \label{eq:aggregation_unified}
    \begin{aligned} 
      \mathbf{m}_{u\to u} + \sum_{v\in \mathcal{A}}\mathbf{m}_{v\to u}, 
    \end{aligned}
\end{equation}
where  $\mathcal{A}$ denotes the aggregation set, which is the set of $u$'s neighbouring nodes $\mathcal{N}(u)$ for the message-passing update and is the rest of the graph nodes for the attention update.
The subscripts are omitted for simplicity. 

The aggregation method of Eq. (\ref{eq:aggregation_unified})  is widely used in current message-passing graph neural networks and Transformer-based models and has achieved considerable success in  empirical evaluations across diverse tasks.
Nonetheless, an issue with this aggregation method is that it is unknown if the primary information in the aggregated representation is from a node itself or from the aggregation set.
Consequently, there exists information lost in the generated  embeddings at each layer, 
and the information lost could be accumulated and  become more serious when more layers are used in the model. 

In this work, we address the issue of information lost in the aggregation method by introducing a differential encoding layer.
First, we compute the differential representation between the information from the aggregation set $\mathcal{A}$ and that from a node itself.
Then we use the differential encoding layer to encode the differential representation and combine the obtained differential encoding with the original aggregated information as the representation aggregated at node $u$:  
\begin{equation} \label{eq:dis}
    \begin{aligned} 
    \mathbf{m}_{u} &=\mathbf{m}_{u\to u} + \sum_{v\in \mathcal{A}}\mathbf{m}_{v\to u} + \\ 
    & \hspace{20pt} \text{diff\_enc}(\sum_{v\in \mathcal{A}}\mathbf{m}^{(k)}_{v\to u} - \mathbf{m}^{(k)}_{u\to u}) \\
    &= \mathbf{m}_{\mathcal{A}+\{u\}} + \text{diff\_enc}(\sum_{v\in \mathcal{A}}\mathbf{m}^{(k)}_{v\to u} - \mathbf{m}^{(k)}_{u\to u}),\\
    \end{aligned}
\end{equation}
where $\text{diff\_enc}$ denotes the differential encoding layer, which is implemented using a position-wise feed-forward network (FFN) in this work.

The proposed differential encoding method draws inspiration from early feature encoding methods.
In image representation, the VLAD  method \cite{jegou2010aggregating} encodes the differential representations between local image features and vectors in a codebook into a compact representation.
This method can be viewed as a simplified form of the Fisher kernel representation approach.
It showed excellent performance in terms of both efficiency and accuracy for tasks including image classification and image retrieval.
Also in vector quantization \cite{jegou2010product},
encoding residual vectors was shown to be more effective than encoding the original vectors.
In deep learning, ResNets \cite{he2016deep} employs residual connections to enable layers to learn the differential representations between their output and input.

Unlike residual representation learning in ResNets, our differential encoding method explicitly encodes the differential representation between the information from the aggregation set and that from a node itself. 
It can be incorporated into current message-passing graph neural networks and Transformer-based models.
By combining the differential encoding, the representational ability of generated embeddings is improved at each message-passing update and the attention update, and therefore the overall graph representation learning performance is improved.

\subsection{ Model Architecture}

Our model architecture for learning over graphs is a hybrid of message-passing and global attention architecture (see Figure \ref{fig:overview}), drawing inspiration from the work of Rampavsek et al. \cite{rampavsek2022recipe}.
A layer of the model encoder consists of a message-passing branch and a parallel global attention branch. 
The message-passing branch generates representations by aggregating features from a node's local neighbourhood, while the global self-attention branch generates representations by aggregating information from all graph nodes.
The representations generated by the two branches are fused together and then fed to an FFN to generate the updated node embeddings.
This update process can be described as follows:

\begin{equation} \label{eq:loss_int}
    \begin{aligned}
    \bar{\mathbf{h}}_{u}^{(k)} &= \text{BN}( \mathbf{h\_{local}}_u^{(k)}) + \text{BN}(\mathbf{h\_{global}}_u^{(k)}) + \mathbf{h}_{u}^{(k-1)} \\
    \mathbf{h}_{u}^{(k)} &= \text{FFN}(\bar{\mathbf{h}}_{u}^{(k)}) + \bar{\mathbf{h}}_{u}^{(k)}, 
    \end{aligned}
\end{equation}
where $\mathbf{h\_{local}}_u^{(k)}$ and $\mathbf{h\_{global}}_u^{(k)}$ are the representations generated by the message-passing  branch and the global attention branch respectively, and $\text{BN}$ denotes the batch normalization layer.
Residual connections are also used in the update process.

\begin{table*}[th]
\caption{Details of the seven benchmark datasets used in the experiments.} 
\centering  
\begin{tabular}{ l|c|cccccc }
\toprule
Dataset  &Graphs &Avg. Nodes/graph &\#Training &\#Validation &\#Test &\#Categories & Task\\

\midrule[.5pt]

MNIST-SP   &70K &40-75  &55,000 &5000 &10,000 & 10 &\multirow{2}{*}{Superpixel graph classification}\\
CIFAR10-SP &60K &85-150 &45,000 &5000 &10,000 & 10 &\\
\midrule[.5pt]
PascalVOC-SP & 11,355 &479.40 & 8,489 &1,428 &1,429 &20 & \multirow{2}{*}{Node classification} \\
COCO-SP & 11,355 &476.88 & 113,286 &5,000 &5,000 &81 &  \\

\midrule[.5pt]
PCQM-Contact &529,434 &30.14 &90\% &5\% &5\% &-- & Long-range link prediction \\
\midrule[.5pt]

Peptides-Func &15,535 &150.90 &70\% &15\% &15\% &10 &Multi-label graph classification\\

\midrule[.5pt]

OGBG-PPA &158,100 &243.4 &70\%  &20\%  &10\%  &37 &Graph classification \\

\bottomrule

\end{tabular}

\label{table:datasets}
\end{table*}

\textbf{Message-passing Update.} 
The message-passing branch uses an MPNN layer to generate the representation for every node $u\in \mathcal{V}$ by aggregating features from its local graph neighbourhood.
We apply our differential encoding method in aggregating the local neighbourhood features.
With differential encoding, the message-passing update can be described as follows:

\begin{equation} \label{eq:loss_int}
    \begin{aligned} 
    \mathbf{h\_local}_{u}^{(k)} &= \text{Update}^{(k)}(\text{Aggregate}^{(k)}(\\
    & \hspace{50pt} \{ {\mathbf{h}_u^{(k-1)}} \} \cup \{{\mathbf{h}_v^{(k-1)}}, \forall v \in \mathcal{N}(u)\} )) \\ 
    &= \text{Update}^{(k)} (\sum_{v\in\mathcal{N}{(u)}+\{u\}}\mathbf{m}_{v\to u} +  \\
    & \hspace{50pt}\text{diff\_enc}_l^{(k)} (\sum_{v\in\mathcal{N}{(u)}}\mathbf{m}_{v\to u}^{(k)} - \mathbf{m}_{u\to u}^{(k)} ) ). \\
    \end{aligned}
\end{equation}
The MPNN layer can be any of the graph neural network layers, such as GCN and GatedGCN.
We validate our differential encoding method on different types of  MPNN layers to show that it is the general method to improve the performance of the message-passing update.

\textbf{Global Attention.} 
The global attention branch uses a multi-head attention function to aggregate features from all graph nodes.
The multi-head attention function generates  representations for all graph nodes as follows:

\begin{equation} \label{eq:loss_int}
    \begin{aligned} 
    \left ( \begin{matrix}  \mathbf{h\_global}_{1}^{(k)} \\
    \vdots \\
    \mathbf{h\_global}_{|\mathcal{V}|}^{(k)} \\
    \end{matrix} \right) &=  \text{MHA}^{(k)}(G,\{\mathbf{h}_v^{(k-1)},v\in \mathcal{V}\}) \\
     & = \text{Concat}(\mathbf{H}_1^{(k)}, ...,\mathbf{H}_{N_h}^{(k)}) \mathbf{W}_{MHA}^{(k)}, \\
    \end{aligned}
\end{equation}
where $\text{MHA}$  denotes the multi-head attention function,  Concat denotes the concatenation operation, $\mathbf{H}^{(k)}_i$ is the output of the $i$-th attention head, $N_h$ is the number of attention heads  and $\mathbf{W}_{MHA}^{(k)}$ is a trainable weight matrix.  
The differential encoding is applied in each attention head,  generating the output $\mathbf{H}^{(k)}_i$  as follows:
\begin{equation} \label{eq:diff_transformer}
    \begin{aligned} 
    \mathbf{H}_i^{(k)} &= \text{Attention}(\mathbf{Q}_i^{(k)},\mathbf{K}_i^{(k)},\mathbf{V}_i^{(k)}) \\
    &= \text{Softmax}(\frac{\mathbf{Q}_i^{(k)}{\mathbf{K}_i^{(k)}}^{T}}{\sqrt{d_q^{(k)}}})\mathbf{V}_i^{(k)} +  \\
    & \hspace{20pt}\text{diff\_enc}_{g,i}^{(k)}( \text{Softmax}(\frac{\mathbf{Q}_i^{(k)}{\mathbf{K}_i^{(k)}}^T}{\sqrt{d_q^{(k)}}})\mathbf{V}_i^{(k)} - \\
    & \hspace{60pt} 2 \mathbf{I}_{d_{i}}^{(k)}\text{Softmax}(\frac{\mathbf{Q}_i^{(k)}{\mathbf{K}_i^{(k)}}^T}{\sqrt{d_q^{(k)}}})\mathbf{V}_i^{(k)}),
    \end{aligned}
\end{equation}
where $\mathbf{Q}_i^{(k)}$, $\mathbf{K}_i^{(k)}$, $\mathbf{V}_i^{(k)}$ are transformed from the input node embeddings, and $\mathbf{I}_{d_{i}}^{(k)}$ is an identity matrix.
The differential encoding method can be efficiently integrated in the current attention mechanism.

Combining a global attention update enables the model to capture information from far reaches of the graph.
Otherwise, the model would generate over-smoothed embeddings after several layers due to the over-smoothing issue and, therefore, could not be able to capture long-term dependencies of the graph using deeper layers.

\textbf{Position-wise FFN.} 
As in the Transformer architecture, the position-wise $\text{FFN}$ consists of two fully connected layers with a rectified linear activation (ReLU) function in between:
\begin{equation} \label{eq:loss_int}
    \begin{aligned} 
    \text{FFN}(x) = \text{fc}_2(\text{ReLU}(\text{fc}_1(x))),
    \end{aligned}
\end{equation}
where $\text{fc}_1$ and $\text{fc}_2$ are the fully connected layers.

\textbf{Graph Representation.} 
The graph representation for $\mathcal{G}$ is computed using  a Readout function that aggregates  the embeddings of all nodes generated by the last encoder layer, resulting in a fixed-size representation:
\begin{equation} \label{eq:loss_int}
    \begin{aligned} 
    \mathbf{h}_{\mathcal{G}} = \text{Readout}(\{ \mathbf{h}_u^{(K)}, u\in \mathcal{V}\}).
    \end{aligned}
\end{equation}
The  graph presentation $\mathbf{h}_{\mathcal{G}}$ can be used for graph-level tasks, such as graph classification and graph regression.
For node-level tasks, the node embeddings generated by the last encoder layer, i.e., $\{ \mathbf{h}_v^{(K)}, v\in \mathcal{V}\}$ are used.
The model is optimized using standard stochastic gradient descent by minimizing a loss function defined according to the target task.

\fi

\iftrue
\section{Experiments}

We experimentally evaluate our model on different tasks on various datasets,  demonstrating that the proposed differential encoding method is a general method to improve the graph representation learning performance.

\begin{table}[th] 
\caption{Results on MNIST and CIFAR10 on the superpixel graph classification task. The differential encoding is applied to the base model GPS. The second-best results are shown in pink.}
\centering
\begin{tabular}{lcccc}
\toprule
\multirow{2}{*}{Model} &MNIST-SP &CIFAR10-SP \\
                        & (Accuracy) &(Accuracy)\\
\midrule

GCN \cite{kipf2016semi}  &90.705$\pm$0.218 &55.710$\pm$0.381 \\
MoNet \cite{monti2017geometric} &90.805$\pm$0.032 &54.655$\pm$0.518 \\
GraphSAGE \cite{hamilton2017inductive} &97.312$\pm$0.097 &65.767$\pm$0.308 \\
GIN   \cite{xu2019powerful}     &96.485$\pm$0.252 &55.255$\pm$1.527\\
GCNII \cite{chen2020simple}   &90.667$\pm$0.143   &56.081$\pm$0.198  \\


PNA \cite{corso2020principal}  &97.94$\pm$0.12 &70.35$\pm$0.63\\

DGN   \cite{beaini2021directional}    &-- &72.838$\pm$0.417\\
Cy2C-GNNs \cite{choi2022cycle} &97.772$\pm$0.001 &64.285$\pm$0.005\\
ARGNP \cite{cai2022automatic} &-- &73.90$\pm$0.15\\

CRaWl  \cite{toenshoff2021graph}  &97.944$\pm$0.050 &69.013$\pm$0.259 \\
GIN-AK+ \cite{zhao2021stars}  &-- &72.19$\pm$0.13\\
3WLGNN \cite{maron2019provably}  &95.075$\pm$0.961 &59.175$\pm$1.593 \\

EGT  \cite{hussain2022global}     &98.173$\pm$0.087 &68.702$\pm$0.409 \\
GatedGCN \cite{bresson2017residual}&97.340$\pm$0.143 &67.312$\pm$0.311 \\
GatedGCN + SSFG \cite{zhang2022ssfg} &97.985$\pm$0.032 &71.938$\pm$0.190 \\
EdgeGCN \cite{zhang2023learning} &98.432$\pm$0.059 &76.127$\pm$0.402 \\
\hline

Exphormer \cite{shirzad2023exphormer} &\color{magenta}{98.550$\pm$0.039} &74.754$\pm$0.194 \\
TIGT \cite{choi2024topology} &98.230$\pm$0.133 &73.955$\pm$0.360 \\
GRIT \cite{ma2023graph} &98.108$\pm$0.111 &\color{magenta}{76.468$\pm$0.881}\\
\midrule
GPS \cite{rampavsek2022recipe}  (base model)    &98.051$\pm$0.126 &72.298$\pm$0.356 \\
\textbf{Ours}  &\textbf{98.558$\pm$0.057} &\textbf{79.067$\pm$0.269}\\
\bottomrule
\end{tabular}
\label{tab:cifar_minist}
\end{table}

\begin{table}[th]
\caption{Results on PascalVOC-SP and COCO-SP on the superpixel graph classification task.  The second-best results are shown in pink.}
\centering  %
\begin{tabular}{l cc }
\toprule
\multirow{2}{*}{Model} &PascalVOC-SP  &COCO-SP\\
                       &(F1)          &(F1) \\
\midrule

GCN \cite{kipf2016semi} & 0.1268$\pm$0.0060  &0.0841$\pm$0.0010 \\
GINE \cite{hu2019strategies}  & 0.1265$\pm$0.0076  &0.1339$\pm$0.0044\\
GCNII \cite{chen2020simple} &0.1698$\pm$0.0080 &0.1404$\pm$0.0011 \\
GatedGCN \cite{bresson2017residual} &0.2873$\pm$0.0219 &0.2641$\pm$0.0045 \\

GatedGCN + RWSE \cite{rampavsek2022recipe}  &0.2860$\pm$0.0085 &0.2574$\pm$0.0034 \\

\hline
Transformer + LapPE \cite{dwivedi2022long}  &0.2694$\pm$0.0098  &0.2618$\pm$0.0031\\

SAN + LapPE \cite{dwivedi2022long}  &0.3230$\pm$0.0039 &0.2592$\pm$0.0158 \\

SAN + RWSE \cite{dwivedi2022long}  &0.3216$\pm$0.0027  &0.2434$\pm$0.0156 \\

Exphormer \cite{shirzad2023exphormer} &\color{magenta}0.3975$\pm$0.0037 &\color{magenta}0.3455$\pm$0.0009 \\

\midrule

GPS \cite{rampavsek2022recipe} &0.3748$\pm$0.0109  &0.3412$\pm$0.0044 \\
\textbf{Ours} &\textbf{0.4242$\pm$0.0011}  &\textbf{0.3567$\pm$0.0026}\\

\bottomrule
\end{tabular}
\label{tab:pascal_coco}
\end{table}

\begin{table*}[th]
\caption{Results on PCQM-Contact on the link prediction task.  The second-best results are shown in pink.}
\centering  
\begin{tabular}{l cccc }
\toprule
Model &Hits@1 &Hit@3 &Hit@10 &MRR ($\uparrow$)\\
\midrule

GCN &0.1321$\pm$0.0007 &0.3791$\pm$0.0004 &0.8256$\pm$0.0006 &0.3234$\pm$0.0006\\
GCNII &0.1325$\pm$0.0009 &0.3607$\pm$0.0003 &0.8116$\pm$0.0009 &0.3161$\pm$0.0004\\
GINE &0.1337$\pm$0.0013 &0.3642$\pm$0.0043 &0.8147$\pm$0.0062 &0.3180$\pm$0.0027 \\
GatedGCN &0.1279$\pm$0.0018 &0.3783$\pm$0.0004 &0.8433$\pm$0.0011 &0.3218$\pm$0.0011 \\
GatedGCN+RWSE  &0.1288$\pm$0.0013 &0.3808$\pm$0.0006 &0.8517$\pm$0.0005 &0.3242$\pm$0.0008 \\
\hline
Transformer+LapPE &0.1221$\pm$0.0011 &0.3679$\pm$0.0033 &0.8517$\pm$0.0039 &0.3174$\pm$0.0020\\
SAN+LapPE &0.1355$\pm$0.0017 &0.4004$\pm$0.0021 &0.8478$\pm$0.0044 &0.3350$\pm$0.0003\\
SAN+RWSE &0.1312±0.0016 &0.4030$\pm$0.0008 &0.8550$\pm$0.0024 &0.3341$\pm$0.0006 \\
Graph Diffuser \cite{glickman2023diffusing}  &0.1369$\pm$0.0012	&\color{magenta}0.4053$\pm$0.0011	&\textbf{0.8592$\pm$0.0007} 	&0.3388$\pm$0.0011 \\

\midrule

GPS \cite{rampavsek2022recipe} &\color{magenta}0.1471$\pm$0.0008 &0.3937$\pm$0.0019 &0.8526$\pm$0.0014 &\color{magenta}0.3337$\pm$0.0006  \\
\textbf{Ours} &\textbf{0.1490$\pm$0.0004} &\textbf{0.4134$\pm$0.0010} &\color{magenta}0.8557$\pm$0.0003 &\textbf{0.3459$\pm$0.0005}\\

\bottomrule

\end{tabular}
\label{tab:pcqm-contact}
\end{table*}

\begin{table}[th]
\caption{Results on OGBG-PPA on the graph classification task.  The second-best results are shown in pink.}
\centering  
\begin{tabular}{l c }
\toprule
\multirow{2}{*}{Model} &OGBG-PPA \\
                       &(Accuracy) \\
\midrule
	
GCN \cite{kipf2016semi} &0.6839$\pm$0.0084 \\
GCN + virtual node &0.6857$\pm$0.0061 \\
	
GIN \cite{xu2019powerful} &0.6892$\pm$0.0100 \\
GIN + virtual node &0.7037$\pm$0.0107 \\
DeeperGCN          &0.7712$\pm$0.0071 \\
ExpC \cite{yang2022breaking}          &0.7976$\pm$0.0072   \\
K-Subtree SAT &0.7522$\pm$0.0056 \\
\midrule

GPS \cite{rampavsek2022recipe}  &\color{magenta}0.8015$\pm$0.0033  \\
\textbf{Ours} &\textbf{0.8096$\pm$0.0029} \\

\bottomrule

\end{tabular}
\label{tab:ppa}
\end{table}

\begin{table}[th]
\caption{Results on Pepti-func on the multi-label graph classification task.  The second-best results are shown in pink.}
\centering  
\begin{tabular}{l cc }
\toprule
Model &AP ($\uparrow$)  \\
\midrule

GCN &0.5930$\pm$0.0023\\
GINE &0.5498$\pm$0.0079 \\

GCNII \cite{chen2020simple} &0.5543$\pm$0.0078 \\
GatedGCN &0.5864$\pm$0.0077\\
Gated + RWSE  & 0.6069$\pm$0.0035 \\
\hline
Transformer+LapPE &0.6326$\pm$0.0126\\
SAN+LapPE &0.6384$\pm$0.0121\\
SAN+RWSE &0.6439$\pm$0.0075 \\

Exphormer \cite{shirzad2023exphormer} &0.6527$\pm$0.0043 \\
\midrule

GPS \cite{rampavsek2022recipe} &\color{magenta}0.6535$\pm$0.0041  \\
Ours &\textbf{0.6608$\pm$0.0046} \\

\bottomrule

\end{tabular}
\label{tab:peptifunc}
\end{table}

\begin{table*}[th]
\caption{Ablation study: Effect of differential encoding on the message-passing update and global attention update. The use of the differential encoding method consistently improves the performance of   message-passing graph neural networks on the seven datasets, and it improves the performance of the global attention update on most of the datasets. } 
\centering  
\begin{tabular}{l cccc }
\toprule
\multirow{2}{*}{Model} &MNIST-SP &CIFAR10-SP &PascalVOC-SP &COCO-SP\\
                       &(Accuracy (\%)) &(Accuracy (\%)) &(F1) &(F1) \\
\midrule[.6pt]

MPNN only &97.989$\pm$0.118 &76.823$\pm$0.219 &0.3579$\pm$0.0047  &0.2446$\pm$0.0027\\
MPNN + differential encoding &\textbf{98.285$\pm$0.080}  &\textbf{77.820$\pm$0.091}  &\textbf{0.3666$\pm$0.0035}  &\textbf{0.2500$\pm$0.0012} \\
\hline
Global attention only & 97.564$\pm$0.146 &45.237$\pm$1.349  &0.2748$\pm$0.0052 &0.2535$\pm$0.0047\\
Global attention + differential encoding  &\textbf{97.912$\pm$0.062} &\textbf{64.468$\pm$0.537}  &\textbf{0.2995$\pm$0.0013}   &\textbf{0.2608$\pm$0.0013} \\
\midrule[.6pt]

Full Model &98.528$\pm$0.057 &79.067$\pm$0.269 &0.4242$\pm$0.0011  &0.3567$\pm$0.0026 \\

\bottomrule
\vspace{-8pt}
\end{tabular}

\begin{tabular}{l ccc }

\toprule
\multirow{2}{*}{Model} &PCQM-Contact &Peptide-func &OGBG-PPA \\
                       &(MRR) &(AP) &(Accuracy)\\
\midrule[.6pt]

MPNN only &0.3306$\pm$0.0004 &0.6131$\pm$0.0043 &0.7995$\pm$0.0042   \\
MPNN + differential encoding &\textbf{0.3406$\pm$0.0002} &\textbf{0.6201$\pm$0.0025}  &\textbf{0.8059$\pm$0.0023}\\
\hline
Global attention only &0.3276$\pm$0.0012  &0.6207$\pm$0.0088 &0.0948$\pm$0.0000 \\
Global attention + differential encoding &\textbf{0.3277$\pm$0.0018}  &\textbf{0.6355$\pm$0.0067} &\textbf{0.0948$\pm$0.0000} \\
\midrule[.6pt]

Full Model &0.3459$\pm$0.0005  &0.6608$\pm$0.0046 &0.8096$\pm$0.0029   \\

\bottomrule

\end{tabular}
\label{table:ablation}
\end{table*}

\subsection{Datasets and Setup}
\textbf{Datasets.}
The experiments are conducted on four graph tasks on the following seven benchmark datasets. 
\begin{itemize}
  \item \textbf{MNIST-SP} and \textbf{CIFAR10-SP} \cite{dwivedi2023benchmarking}.
      The two datasets consists of superpixel graphs  extracted from the images in the  MNIST  dataset \cite{lecun1998gradient} and CIFAR10 dataset \cite{krizhevsky2009learning} using the SLIC method \cite{achanta2012slic}.
      The superpixels represent a small region of homogeneous intensity in the original images.
      The two datasets are used for evaluation on the graph classification task.
      
  \item \textbf{PascalVOC-SP} and \textbf{COCO-SP} \cite{dwivedi2022long}. The two datasets are from the long-range graph benchmark (LRGB) datasets, which were introduced to validate a model's performance for capturing long-term dependencies. PascalVOC-SP and COCO-SP are also superpixel graph datasets, wherein the superpixel graphs are extracted using the SLIC method \cite{achanta2012slic} from the images in the Pascal VOC 2011 dataset \cite{everingham2010pascal} and COCO dataset \cite{lin2014microsoft} respectively.
    The two datasets are  used for evaluation on the node classification task.
    
  \item \textbf{PCQM-Contact} \cite{dwivedi2022long}. 
  The PCQM-Contact dataset is also from the LRGB datasets. 
   This dataset contains 529,434 graphs with approximately 15 million nodes.
  The graphs are from the PCQM4M training dataset \cite{hu2020open}. 
   Each graph represents a molecular with explicit hydrogens.
  The task on this dataset is to predict if pairs  nodes in a molecule graph from a distance (more than 5 hops away) will be contacting with each other in the 3D space.

   \item \textbf{Peptides-func} \cite{dwivedi2022long}.  This Peptides-func dataset is also from the LRGB datasets. This datasets consists of peptides molecular graphs, in which the nodes  represent heavy (nonhydrogen) atoms of the peptides and the edges represent the bonds between these atoms. These graphs are categorized into 10 classes based on the peptide functions, e.g., antibacterial, antiviral, cell-cell communication. This dataset is used for evaluation on the multi-label graph classification task.

   \item  \textbf{OGBG-PPA} \cite{hu2020open}
   consists of protein-protein association (PPA) networks derived from 1581 species categorized into 37 taxonomic groups. 
   The nodes represent proteins, and the edges encode the normalized level of 7 different associations between two proteins. The task on this dataset is to classify which of the 37 groups a PPA network originates from.
   
\end{itemize}

The details of the seven benchmark datasets are reported in Table \ref{table:datasets}.

\textbf{Evaluation metrics.}
Following the work of Dwivedi et al. \cite{dwivedi2023benchmarking} and Rampavsek et al. \cite{rampavsek2022recipe}, the following metrics are utilized for performance evaluation for different datasets.
\begin{itemize}

   \item \textbf{Accuracy} is used for evaluating the model performance  on the superpixel classification task on PascalVOC-SP and COCO-SP and on the PPA network classification task on OGBG-PPA.
  
  \item  \textbf{F1 score}. The results on PascalVOC-SP and COCO-SP on the node classification task is evaluated using the macro weighted F1 score.

   \item \textbf{Mean Reciprocal Rank (MRR)}. The results on link prediction on the PCQM-Contact dataset is evaluated using MRR (a.k.a. inverse harmonic mean rank \cite{hoyt2022unified}),  Hits@1, Hits@3 and Hits@10 are also reported.

   \item \textbf{Average Precision (AP)}. The results on the multi-label graph classification task on Peptides-func is evaluated using unweighted mean AP.

\end{itemize}

\textbf{Implementation Details.}
We closely followed the experimental setup in Rampasek et al. \cite{rampavsek2022recipe} in our implementation. 
We used the same train/validation/test split of each dataset and report the  mean and standard deviation over 10 runs.
The AdamW algorithm \cite{loshchilov2017decoupled} with a cosine learning rate schedule with warmup is used for training  the models.
The values of $\beta_1$, $\beta_2$ and $\epsilon$ in the AdamW algorithm are set to 0.9, 0.999 and $10^{-8}$, respectively.
The base learning rate and training epochs are different for different datasets.
The setups for the model hyper-parameters are also different for different datasets.
The details can be found  in the appendix section.

\subsection{Experimental Results}

\textbf{MNIST-SP and CIFAR10-SP.}
We used GPS as the base model, wherein GateGCN is employed for the message-passing update. 
The results on the two datasets are reported in Table \ref{tab:cifar_minist}.
Our model achieves an accuracy of 98.558\% and 79.067\% on the two datasets respectively.
As compared to the based model GPS, the use of our differential encoding method improves the accuracy by 0.477\% and 6.769\% on MNIST-SP and CIFAR10-SP respectively.
EdgeGCN achieves the best performance among the baseline graph neural network models.
EdgeGCN aggregates local edge features and also employs SSFG in generating node embeddings.
Our model achieves 2.94\% improved accuracy on CIFAR10-SP compared to EdgeGCN.
Our model also outperforms the  baseline Transformer-based models, i.e., Exphormer, TIGT and GRIT on the two datasets.
GRIT \cite{ma2023graph} achieves best performance among the baseline models on CIFAR10-SP.
When compared to GRIT, our model improves the performance by 2.599\% on this dataset.
The quantitative results suggest that combining global attention with message-passing generates improved graph representations compared the use of a single approach.
To the best of our knowledge, our model achieves the state-of-the-art results on the two datasets.

\textbf{PascalVOC-SP and COCO-SP.}
The results on the two datasets are reported in Table \ref{tab:pascal_coco}.
As with the base model GPS, our model combines the GatedGCN update and global attention update in an encoder layer. 
Overall, our model achieves an F1 score of 0.4242 and 0.3567  on the two datasets, respectively, outperforming all the baseline graph neural network-based models (GCN, GINE, GCNII and GatedGCN) and  Transformer-based models (SAN and Exphormer).
Compared to GPS, the use of our differential encoding method improves the F1 score from 0.3748 to 0.4242 and from 0.3412 to 0.3567 on the two datasets respectively. 
Notably, the use of differential encoding yields a 0.0494 performance gain  on PascalVOC-SP, which is a relative 13.18\% improvement.
Exphormer \cite{shirzad2023exphormer} is the best modes among the baselines.
It  is a spare attention-based Transformer architecture which  uses virtual nodes and  augmented graphs.
When compared to Exphormer, our model achieves  a relative improvement of 6.7\% and 3.2\% on the two datasets respectively.


\textbf{PCQM-Contact.} 
Table \ref{tab:pcqm-contact} reports the results on PCQM-Contact on the link prediction task.
Our model achieves an MRR of 0.3459 on this dataset, outperforming all baseline graph neural network-based models and Transformer-based models.
As compared to the base model GPS, the use of differential encoding improves the MRR from 0.3337 to 0.3459, which is a relative 3.7\% performance improvement.
Our model also improves the performance in terms of Hit@1, Hit@3 and Hit@10.
Once again, our models results in new  state-of-the-art result for graph representation learning on this dataset. 

\textbf{OGBG-PPA.}  Table \ref{tab:ppa} reports the results on OGBG-PPA on the PPA network classification task. 
Our model achieves an accuracy of 80.96\% on this dataset, outperforming all the baseline models.
Following the work of Rampavsek et al. \cite{rampavsek2022recipe}, the models that require pretraining on another dataset or use an ensemble prediction approach are not include in the baselines.
Our model outperforms conventional graph neural network models, i.e., GCN and GIN, by a large margin.
Compared to the base model GPS, the use of differential encoding improve the classification accuracy by 0.81\%.

\textbf{Peptides-func.} 
Table \ref{tab:peptifunc} reports the results on Peptides-func on the multi-label graph classification task. 
The use of differential encoding improve the mean average precision of the base model GPS from 65.35\% to 66.08\%.
Our model outperforms all the baseline graph neural network-based models, i.e., GCN, GINE, GCNII and GatedGCN, and  Transformer-based models, i.e., Transformer with Laplacian positional encoding, SAN and Exphormer.

From Table \ref{tab:cifar_minist} to Table \ref{tab:peptifunc}, we  see that 
integrating differential encoding yields a lower variance in the results on six of the seven datasets compared to the base model.
This suggests that our differential encoding  method helps improve the numerical stability of optimization.

\begin{table}[th]
\caption{Effect of our differential encoding method on other graph neural network models.}
\centering  %
\begin{tabular}{lccc }
\toprule
\multirow{2}{*}{Model}  &MNIST-SP &CIFAR10-SP  \\
                        &(Accuracy) &(Accuracy) \\
\midrule

GCN   &90.120$\pm$0.145 &54.142$\pm$0.394\\
GCN + differential enc.   &\textbf{94.728$\pm$0.137} &\textbf{62.020$\pm$0.362}\\
                          &(4.608\%$\uparrow$)  &(7.878\%$\uparrow$)  \\
\hline

\hline
GAT   &95.535$\pm$0.205 & 64.223$\pm$0.455  \\
GAT + differential enc.   &\textbf{97.163$\pm$0.154} &\textbf{67.373$\pm$0.415}\\
                          &(1.628\%$\uparrow$) &(3.120\%$\uparrow$) \\

\bottomrule

\end{tabular}
\label{tab:diffenc_gcngat}
\end{table}

\textbf{Ablation Study.}
We conducted ablations of our model to demonstrate the effect of differential encoding on the message-passing update and global attention update.
The ablation study results are shown in Table \ref{table:ablation}.
We see that the use of the differential encoding method consistently improves the performance of message-passing graph neural networks on the seven datasets and improves the performance of the global attention update on most of the datasets. 
The results shows the importance of  our differential encoding method in improving the graph representation learning performance.
Notably, the use of our method improve the performance of the global attention update by 19.231\% on CIFAR-SP.
This is highly promising that it could be used in  Transformers for computer vision and sequence learning tasks.

For the quantitative results on the seven benchmark datasets,  
our model uses the GatedGCN layer in the message-passing branch.
We further validated our differential encoding method on two popular graph neural network models, i.e., GCN and GAT, on MNIST-SP and CIFAR10-SP. 
The results are shown in Table \ref{tab:diffenc_gcngat}.
We see that our differential encoding method is also effective in improving the performance of the two models, showing that our differential method is a general method to improve the performance of graph neural networks.

\section{Conclusion}
This paper proposed a differential encoding method  to address the issue of information lost  in the current aggregation approach in the message-passing update and global attention update for graph representation learning. 
The idea of our method is to introduce an encoding layer to encode the differential representation between the information from a node's neighbours (or the rest of graph nodes) and that from the node itself.
The differential encoding is then combined with the original aggregated representation to generate updated node embeddings.
By combining the differential encoding, the relation between the information from a node's neigbhours (or from the rest of graph nodes) and that from the node iteself is integrated in the generated embeddings, and, therefore, the embedding capability of the current message-passing paradigm and global attention mechanism is improved.
We experimentally evaluated our  model on different graph tasks, i.e.,  graph classification, node classification, link prediction and multi-label graph classification, on seven  benchmark datasets, i.e., MNIST-SP, CIRAR10-SP, PascalVOC-SP, COCO-SP, PCQM-Contact, Peptides-func and OGBG-PPA.
We demonstrated that our differential encoding method is a general method to improve the message-passing update and global attention update, advancing the state-of-the-art results for graph representation learning on these datasets.

\fi

\section*{Acknowledgment}
The authors would like to thank the editor and reviewers for reviewing this manuscript.

\bibliographystyle{IEEEtran}
\bibliography{mybib}

\end{document}